%% file: paper.tex
\documentclass[runningheads]{llncs}
\usepackage{graphicx}
\usepackage{amsmath,amssymb} 
\usepackage{color}

\usepackage{float}
\usepackage{algorithm}
\usepackage{algorithmic}
\usepackage{booktabs}

\usepackage{hyperref}
\usepackage[noabbrev,capitalize,nameinlink]{cleveref}
\usepackage{tablefootnote}

\newcommand{\NAS}{NAS}
\newcommand{\NASRL}{\NAS}
\newcommand{\NAScell}{\NAS}

\newcommand{\eat}[1]{}

\begin{document}

\title{Progressive Neural Architecture Search} 

\titlerunning{Progressive Neural Architecture Search}
%
\author{Chenxi Liu\inst{1}\thanks{Work done while an intern at Google.} \and
Barret Zoph\inst{2} \and
Maxim Neumann\inst{2} \and
Jonathon Shlens\inst{2} \and
Wei Hua\inst{2} \and 
Li-Jia Li\inst{2} \and 
Li Fei-Fei\inst{2,3} \and 
Alan Yuille\inst{1} \and 
Jonathan Huang\inst{2} \and 
Kevin Murphy\inst{2}
}
%
\authorrunning{C. Liu et al.}
%

\institute{Johns Hopkins University \and
Google AI \and 
Stanford University}
\maketitle              
\input{abstract}

\input{intro}

\input{related}

\input{space}

\input{methods}

\input{results}

\input{discussion}

\clearpage
\section*{Acknowledgements}
We thank Quoc Le for inspiration, discussion and support; George Dahl for many fruitful discussions; Gabriel Bender, Vijay Vasudevan for the development of much of the critical infrastructure and the larger Google Brain team for the support and discussions. CL also thanks Lingxi Xie for support.

%
%
%
\bibliographystyle{splncs04}
\bibliography{pnas}

\clearpage
\input{suppl}

\end{document}

%% file: abstract.tex
\begin{abstract}
We propose a new method for learning the structure of convolutional neural networks (CNNs) that is more efficient than recent state-of-the-art methods based on reinforcement learning and evolutionary algorithms. Our approach uses a sequential model-based optimization (SMBO) strategy, in which we search for structures in order of increasing complexity, while simultaneously learning a surrogate model to guide the search through structure space. Direct comparison under the same search space shows that our method is up to 5 times more efficient than the RL method of Zoph et al. (2018) in terms of number of models evaluated, and 8 times faster in terms of total compute. The structures we discover in this way achieve state of the art classification accuracies on CIFAR-10 and ImageNet.
\end{abstract}

%% file: intro.tex
\section{Introduction}

There has been a lot of recent interest in automatically learning good neural net architectures.
Some of this work is summarized in \cref{sec:related}, but at a high level,
current techniques
usually fall into one of two categories: evolutionary algorithms 
(see e.g. \cite{DBLP:conf/icml/RealMSSSTLK17,Miikkulainen2017,DBLP:journals/corr/XieY17})
or reinforcement learning
(see e.g., \cite{DBLP:journals/corr/ZophL16,DBLP:journals/corr/ZophVSL17,Zhong2018,Cai2017,DBLP:journals/corr/BakerGNR16}).
When using evolutionary algorithms (EA), each neural network structure is encoded as a string, and random mutations and recombinations of the strings are performed during the search process;
each string (model) is then trained and evaluated on a validation set,
and the top performing models generate  ``children''.
When using reinforcement learning (RL),  the agent performs a sequence of actions, which specifies the structure of the model; this model is then trained and its validation performance is returned as the reward, which is used to update the RNN controller.
Although both EA and RL methods have been able to learn network structures that outperform manually designed architectures, they require significant computational resources.
For example, the RL method in
\cite{DBLP:journals/corr/ZophVSL17} 
trains and evaluates 20,000 neural networks across 500 P100 GPUs over 4 days.

In this paper, we describe a method that is able to learn a CNN which matches previous state of the art in terms of accuracy,
while requiring 5 times fewer model evaluations during the architecture search.
Our starting point is the structured search space
proposed by \cite{DBLP:journals/corr/ZophVSL17}, in which the search algorithm is tasked with searching for a good convolutional ``cell'', as opposed to a full CNN.
A cell contains $B$ ``blocks'', where a block is a combination operator (such as addition) applied to two inputs (tensors), each of which can be transformed (e.g., using convolution) before being combined.
This cell structure is then stacked a certain number of times, depending on the size of the training set, and the desired running time of the final CNN
(see \cref{sec:space} for details).
This modular design also allows easy architecture transfer from one dataset to another,
as we will show in experimental results.

We propose to use  heuristic search to search the space of cell structures,
starting with simple (shallow) models and progressing to complex ones, pruning out unpromising structures as we go.
At iteration $b$ of the algorithm, we have a set of $K$ candidate cells (each of size $b$ blocks), which we train and evaluate on a dataset of interest. Since this process is expensive,
we also learn a model or surrogate function which can predict the performance of a structure
without needing to training it.
We expand the $K$ candidates of size $b$ into $K' \gg K$ children, each of size $b+1$.
We apply our surrogate function to rank all of the $K'$ children,
pick the top $K$, and then train and evaluate them.
We continue in this way until $b=B$, which is the maximum number of blocks we want to use in our cell.
See \cref{sec:method} for details.

Our progressive (simple to complex) approach has several advantages over other techniques that directly search in the space of fully-specified structures. First, the simple structures train faster, so we get some initial results to train the surrogate quickly. Second, we only ask the surrogate to predict the quality of structures that are slightly different
(larger) from the ones it has seen
(c.f., trust-region methods). 
Third, we factorize the search space into a product of smaller search spaces, 
allowing us to potentially search models with many more blocks.
In \cref{sec:results} we show that our approach 
is 5 times more efficient than the RL method of \cite{DBLP:journals/corr/ZophVSL17} in terms of number of models evaluated, and 8 times faster in terms of total compute.
We also show that the structures we discover achieve state of the art classification accuracies on CIFAR-10 and ImageNet.\footnote{The code and checkpoint for the PNAS model trained on ImageNet can be downloaded from the TensorFlow models repository at \url{http://github.com/tensorflow/models/}. Also see \url{https://github.com/chenxi116/PNASNet.TF} and \url{https://github.com/chenxi116/PNASNet.pytorch} for author's reimplementation.}

%% file: related.tex
\section{Related Work}
\label{sec:related}

Our paper is based on the ``neural architecture search'' (NAS) method proposed in
\cite{DBLP:journals/corr/ZophL16,DBLP:journals/corr/ZophVSL17}.
In the original paper  \cite{DBLP:journals/corr/ZophL16},
they
use the REINFORCE algorithm \cite{Williams92} to estimate the parameters of a
recurrent neural network (RNN),
which represents a policy to generate a sequence of symbols (actions) specifying the structure of the CNN; the reward function is the classification accuracy on the validation set of a CNN generated from this sequence.
\cite{DBLP:journals/corr/ZophVSL17} extended this by using a more structured search space,
in which the CNN was defined in terms of a series of stacked ``cells''.
(They also replaced  REINFORCE with 
proximal policy optimization (PPO) \cite{PPO}.)
This method was able to learn CNNs which outperformed almost all previous methods
in terms of accuracy vs speed on image classification (using
CIFAR-10 \cite{krizhevsky2009learning} and ImageNet \cite{DBLP:conf/cvpr/DengDSLL009})
and object detection (using COCO \cite{DBLP:conf/eccv/LinMBHPRDZ14}).

There are several other papers that use RL to learn network structures.
\cite{Zhong2018}  use the same model search space as \NAScell,  but replace policy gradient with Q-learning.
\cite{DBLP:journals/corr/BakerGNR16} also use Q-learning, but without exploiting cell structure.
\cite{Cai2017} use policy gradient to train an RNN, but the actions are now to widen an existing layer, or to deepen the network by adding an extra layer. This requires specifying an initial model and then gradually learning how to transform it.
The same approach, of applying ``network morphisms'' to modify a network,
was used in 
\cite{Elsken2017}, but in the context of hill climbing search, rather than RL.
\cite{ENAS} use parameter sharing among child models to substantially accelerate the search process.

An alternative  to RL is to use evolutionary algorithms (EA; ``neuro-evolution'' \cite{Stanley2017}).
Early work
(e.g., \cite{Stanley2002})
used EA to learn both the structure and the parameters of the network,
but more recent methods,
such as \cite{DBLP:conf/icml/RealMSSSTLK17,Miikkulainen2017,DBLP:journals/corr/XieY17,Liu2017,DBLP:journals/corr/abs-1802-01548},
just use EA to search the structures, and use SGD to estimate the parameters.

RL and EA are local search methods that search through the space of fully-specified graph structures. An alternative approach, which we adopt, is to use heuristic search, in which we search through the space of structures in a progressive way, from simple to complex.
There are several pieces of prior work that explore this approach.
\cite{Negrinho2017} use Monte Carlo Tree Search (MCTS),
but at each node in the search tree, it uses random selection to choose which branch to expand,
which is very inefficient.
Sequential Model Based Optimization (SMBO) \cite{Hutter2011} improves on MCTS by learning a predictive model, which can be used to decide which nodes to expand.
This technique has been applied to neural net structure search in
\cite{Negrinho2017},
but they used a flat CNN search space, rather than our hierarchical cell-based space.
Consequently, their resulting CNNs do not perform very well.
Other related works include
\cite{Mendoza2016}, who focus on MLP rather than CNNs;
\cite{Stanley2002}, who used an incremental approach in the context of evolutionary algorithms; \cite{DBLP:journals/corr/ZophL16} who used a schedule of increasing number of layers;
and \cite{Grosse2012} who 
search through the space of latent factor models specified by a grammar.
Finally, \cite{Cortes2016,Huang2017boosting} grow CNNs sequentially using boosting.

Several other papers learn a surrogate function to predict the performance of a candidate structure, either ``zero shot'' (without training it)
(see e.g., \cite{SMASH}),
or after training it for a small number of epochs and extrapolating the learning curve
(see e.g.,  \cite{Domhan2015,Baker2017acc}).
However, most of these methods have been applied to fixed sized structures, and would not work with our progressive search approach.

%% file: space.tex
\section{Architecture Search Space}
\label{sec:search}
\label{sec:space}

In this section we describe the neural network architecture search space used in our work.
We build on the hierarchical approach
proposed in  \cite{DBLP:journals/corr/ZophVSL17},
in which we first learn a cell structure, and then stack this cell a desired number of times,
in order to create the final CNN.

\subsection{Cell Topologies}

A cell is a fully convolutional network
that maps an $H \times W \times F$ tensor
to another $H' \times W' \times F'$ tensor.
If we use stride 1 convolution, then $H'=H$ and $W'=W$;
if we use stride 2, then $H'=H/2$ and $W'=W/2$.
We employ a common heuristic to double the number of filters (feature maps)
whenever the spatial activation is halved,
so $F'=F$ for stride 1, and $F'=2F$ for stride 2.

The cell can be represented 
by a DAG consisting of $B$ blocks. Each block is a mapping from 2 input
tensors to 1 output tensor.
We can specify a block $b$ in a cell $c$ as a 5-tuple,
$(I_1, I_2, O_1, O_2, C)$,
where
$I_1, I_2 \in \mathcal{I}_b$ specifies the 
inputs to the block,
$O_1, O_2 \in \mathcal{O}$ specifies the operation to apply to input $I_i$,
and $C \in \mathcal{C}$ specifies how to combine $O_1$ and $O_2$ to generate the feature map (tensor) corresponding to  the output of this block,
which we denote by $H_b^c$.

The set of possible inputs,
 $\mathcal{I}_b$, is the
set of all previous blocks in this cell,
$\{H_1^c,\ldots,H^c_{b-1}\}$,
plus the output of the previous cell,
$H_B^{c-1}$,
plus the output of the previous-previous cell,
$H_B^{c-2}$.

The operator space  $\mathcal{O}$
is the following set of 8 functions, each of which
operates on a single tensor\footnote{The depthwise-separable convolutions are in fact two repetitions of ReLU-SepConv-BatchNorm; 1x1 convolutions are also inserted when tensor sizes mismatch.}:

\begin{minipage}{0.5\textwidth}
\renewcommand\labelitemi{$\bullet$}
\footnotesize
\bigbreak
\begin{itemize}
\item 3x3 depthwise-separable convolution
\item 5x5 depthwise-separable convolution
\item 7x7 depthwise-separable convolution
\item 1x7 followed by 7x1 convolution
\end{itemize}
\bigbreak
\end{minipage}
\begin{minipage}{0.5\textwidth}
\renewcommand\labelitemi{$\bullet$}
\footnotesize
\bigbreak
\begin{itemize}
\item identity
\item 3x3 average pooling
\item 3x3 max pooling
\item 3x3 dilated convolution
\end{itemize}
\bigbreak
\end{minipage}
This is less than the 13 operators used in \cite{DBLP:journals/corr/ZophVSL17},
since 
we removed the ones that their RL method discovered were never used.
 
For the space of possible combination operators $\mathcal{C}$,
 \cite{DBLP:journals/corr/ZophVSL17} considerd both
 elementwise addition and concatenation.
However, they discovered that the RL method never chose to use concatenation,
so to reduce our search space, we always use addition as the combination operator.
Thus in our work, a block can be specified by a 4-tuple.

We now quantify the size of the search space to highlight the magnitude of the search problem. 
Let the space of possible structures for the $b$'th block
be $\mathcal{B}_b$;
this has size
$|\mathcal{B}_b|= |\mathcal{I}_b|^2 \times |\mathcal{O}|^2 \times |\mathcal{C}|$,
where 
$|\mathcal{I}_b|=(2+b-1)$,
$|\mathcal{O}|=8$ and $|\mathcal{C}|=1$.
For $b=1$, we have 
$\mathcal{I}_1 = \{H_B^{c-1}, H_B^{c-2}\}$,
which 
are the final outputs of the previous two cells,
so there are $|\mathcal{B}_1|=256$ possible block structures.

If we allow cells of up to $B=5$ blocks,
the total number of cell structures
is given by
$|\mathcal{B}_{1:5}| = 2^2 \times 8^2 \times 3^2 \times 8^2 \times 4^2 \times 8^2 \times 5^2 \times 8^2 \times 6^2 \times 8^2 = 5.6 \times 10^{14}$. 
However, there are certain symmetries in this space that allow us to prune it
to a more reasonable size.
For example, there are only 136 unique cells composed of 1 block.
The total number of unique cells is $\sim 10^{12}$.
This is much smaller than the search space used in 
 \cite{DBLP:journals/corr/ZophVSL17}, which has size
 $10^{28}$,
but it is still an extremely large space to search, and requires efficient 
optimization methods.

\subsection{From Cell to CNN}

\begin{figure}[t]
\centering
\includegraphics[width=0.5\textwidth]{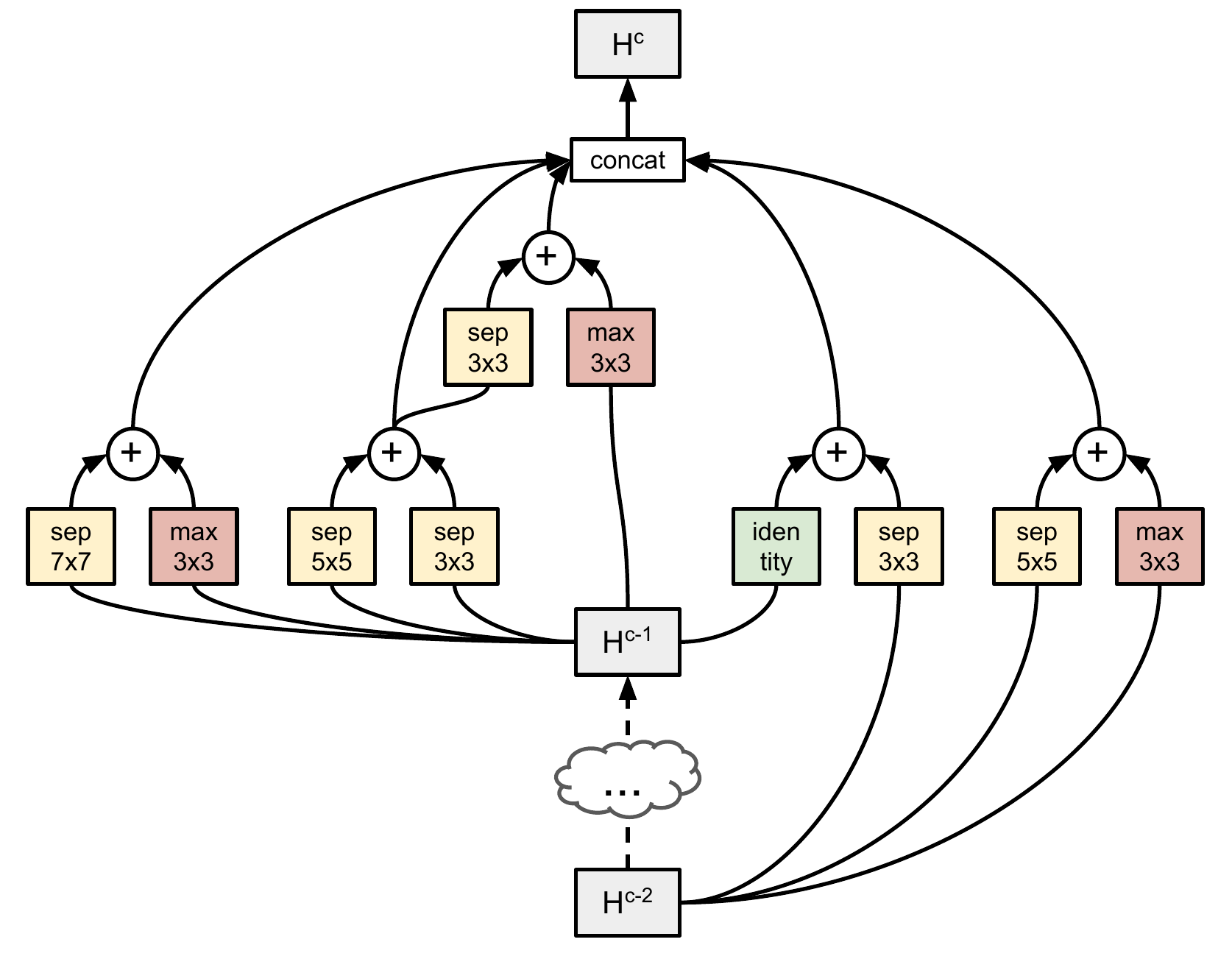}
\qquad
\includegraphics[width=0.3\textwidth]{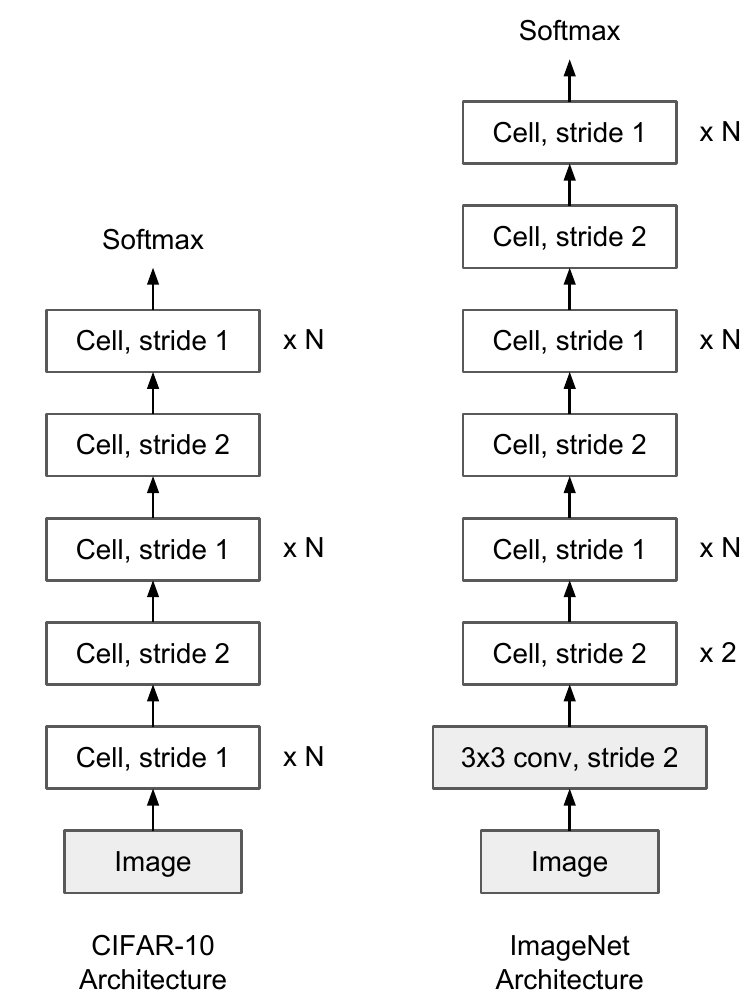}
\caption{
\emph{Left}: 
The best cell structure found by our Progressive Neural Architecture Search, consisting of 5 blocks.
\emph{Right}:
We employ a similar strategy as \cite{DBLP:journals/corr/ZophVSL17} when constructing CNNs from cells on CIFAR-10 and ImageNet. 
Note that we learn a single cell type instead of distinguishing between Normal and Reduction cell.
}
\label{fig:network}
\end{figure}

To evaluate a cell, we have to convert it into a CNN.
To do this, 
we stack a predefined number of copies of the basic cell (with the same structure, but untied weights),
using either stride 1 or stride 2, as shown in 
Figure~\ref{fig:network} (right). 
The number of stride-1 cells between stride-2 cells is then adjusted accordingly with up to $N$ number of repeats.
At the top of the network, we use global average pooling, followed by a softmax classification layer.
We then train the stacked model on the relevant dataset.

In the case of CIFAR-10, we use $32 \times 32$ images.
In the case of ImageNet, we consider two settings,
one with high resolution images of size $331 \times 331$,
and one with smaller images of size $224 \times 224$.
The latter results in less accurate models, but they are faster.
For ImageNet,
we also add an initial $3 \times 3$ convolutional filter layer  with stride 2 at the start of the network, to further reduce the cost.

The overall CNN construction process is identical to  \cite{DBLP:journals/corr/ZophVSL17},
except we only use one cell type (we do not distinguish between Normal and Reduction cells, but instead emulate a Reduction cell by using a Normal cell with stride 2),
and the cell search space is slightly smaller (since we use fewer operators and combiners).

%% file: methods.tex
\section{Method}
\label{sec:methods}
\label{sec:method}

\subsection{Progressive Neural Architecture Search}
\label{sec:pnas}

\begin{algorithm}[t]
\begin{algorithmic}
\STATE {\bfseries Inputs:}
$B$ (max num blocks),
$E$ (max num epochs),
$F$ (num filters in first layer),
$K$ (beam size),
$N$ (num times to unroll cell),
trainSet,
valSet.
\STATE $\mathcal{S}_1$ = $\mathcal{B}_1$ // {\it Set of candidate structures with one block}
\STATE $\mathcal{M}_1$ = cell-to-CNN($\mathcal{S}_1$, $N$, $F$) // {\it Construct CNNs from cell specifications}
\STATE $\mathcal{C}_1$ = train-CNN($\mathcal{M}_1$, $E$, trainSet) //  {\it Train proxy CNNs}
\STATE $\mathcal{A}_1$ = eval-CNN($\mathcal{C}_1$, valSet) //  {\it Validation accuracies}
\STATE $\pi$ = fit($\mathcal{S}_1$, $\mathcal{A}_1)$ // {\it Train the reward predictor from scratch}
\FOR{$b=2:B$}
\STATE $\mathcal{S}'_b$ = expand-cell($\mathcal{S}_{b-1}$) // {\it Expand current candidate cells by one more block}
\STATE $\hat{\mathcal{A}}'_b$ = predict($\mathcal{S}'_b$, $\pi$) // {\it Predict accuracies using reward predictor}
\STATE $\mathcal{S}_b$ = top-K($\mathcal{S}'_b$, $\hat{\mathcal{A}}'_b$, $K$) // {\it Most promising cells according to prediction}
\STATE $\mathcal{M}_b$ = cell-to-CNN($\mathcal{S}_b$, $N$, $F$)
\STATE $\mathcal{C}_b$ = train-CNN($\mathcal{M}_b$, $E$, trainSet) 
\STATE $\mathcal{A}_b$ = eval-CNN($\mathcal{C}_b$, valSet) 
\STATE $\pi$ = update-predictor($\mathcal{S}_b$, $\mathcal{A}_b$, $\pi$) // {\it Finetune reward predictor with new data}
\ENDFOR
\STATE Return top-K($\mathcal{S}_B$, $\mathcal{A}_B$, 1) 
\end{algorithmic}
\caption{Progressive Neural Architecture Search (PNAS).
}
\label{alg:pnas}
\end{algorithm}

\begin{figure}[t]
    \centering
    \begin{minipage}[c]{0.49\textwidth}
    \includegraphics[width=\textwidth]{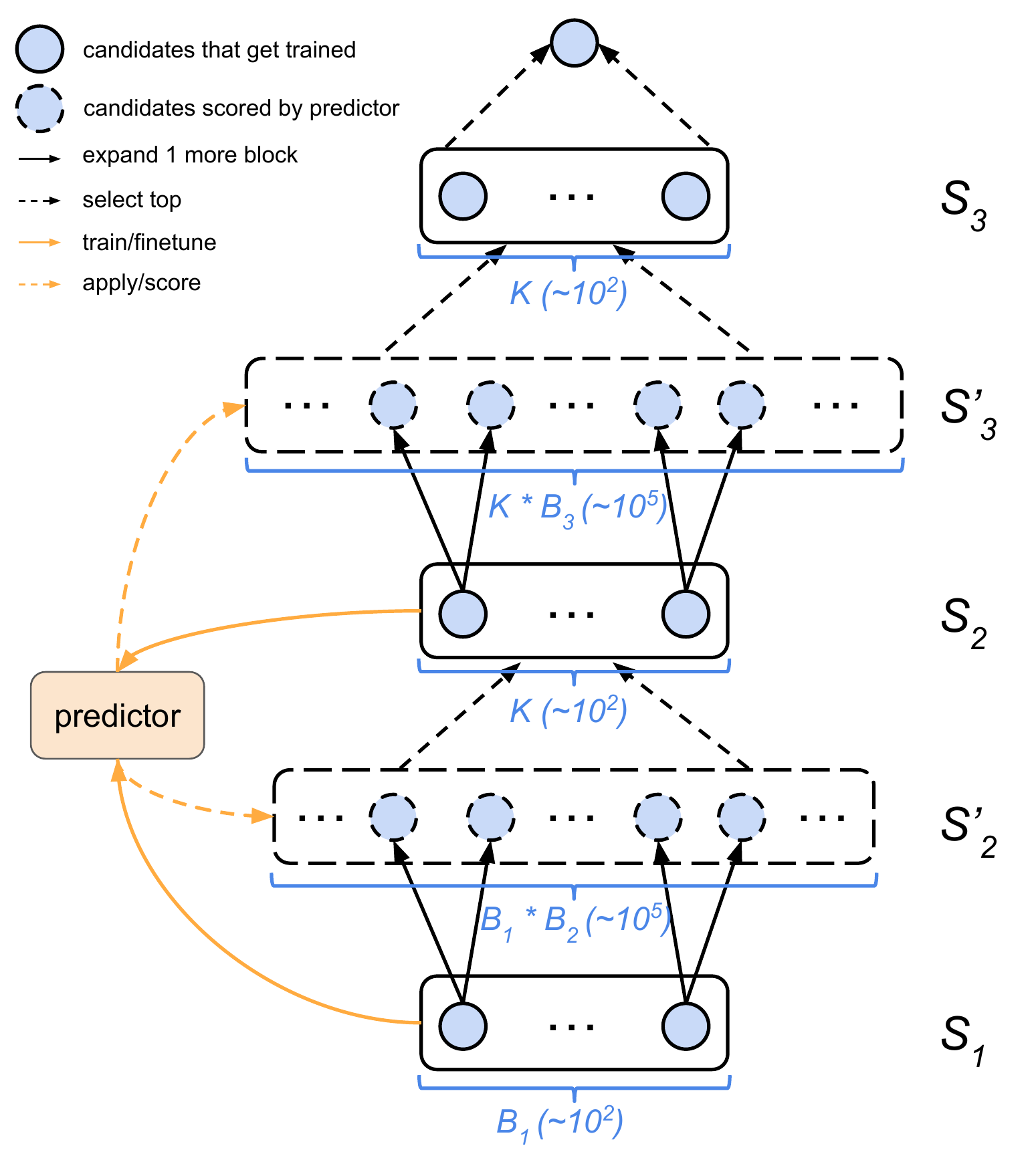}
    \end{minipage} \hfill
    \begin{minipage}[c]{0.5\textwidth}
    \caption{Illustration of the PNAS search procedure when the maximum number of blocks is $B=3$.
    Here $\mathcal{S}_b$ represents the set of candidate cells with $b$ blocks.
    We start by considering all cells with 1 block, $\mathcal{S}_1 = \mathcal{B}_1$;
    we train and evaluate all of these cells, and update the predictor.
    At iteration 2, we expand each of the cells in $\mathcal{S}_1$ 
    to get all cells with 2 blocks, $\mathcal{S}'_2 = \mathcal{B}_{1:2}$;
    we predict their scores, pick the top $K$ to get $\mathcal{S}_2$,
    train and evaluate them, and update the predictor.
    At iteration 3, we expand each of the cells in $\mathcal{S}_2$,  to get a subset of cells with 3 blocks,
    $\mathcal{S}'_3 \subseteq \mathcal{B}_{1:3}$;
     we predict their scores, pick the top $K$ to get $\mathcal{S}_3$, train and evaluate them,
     and return the winner.
 $B_b = |\mathcal{B}_b|$ is the number of possible blocks at level $b$
    and $K$ is the beam size (number of models we
    train and evaluate per level of the search tree).
    }
    \label{fig:PNAStree}
    \end{minipage}
\end{figure}

Many previous approaches directly search in the space of full cells,
or worse,
full CNNs.
For example, \NAScell\ uses a 50-step RNN\footnote{5 symbols per block, times 5 blocks, times 2 for Normal and Reduction cells.
} %
as a controller to generate cell specifications.
In \cite{DBLP:journals/corr/XieY17} a fixed-length binary string encoding of CNN architecture is defined and used in model evolution/mutation.
While this is a more direct approach, we argue that it is difficult to directly navigate in an exponentially large search space, especially at the beginning where there is no knowledge of what makes a good model.

As an alternative, we propose to search the space in a progressive order,
simplest models first.
In particular, we start by constructing all possible cell structures 
from $\mathcal{B}_1$ (i.e., composed of 1 block),
and add them to a queue.
We train and evaluate all the models in the queue (in parallel),
and then expand each one by
adding all of the possible  block structures from $\mathcal{B}_2$;
this gives us a set of
$|\mathcal{B}_1| \times |\mathcal{B}_2| = 256 \times 576 = 147,456$ candidate cells of depth 2.
Since 
we cannot afford to 
train and evaluate  all of these child networks,
 we refer to a learned predictor function
(described in \cref{sec:surrogate});
it is trained based on the measured performance of the cells we have visited so far.
(Our predictor takes negligible time to train and apply.)
We then use the predictor to evaluate all the candidate cells,
and pick the $K$ most promising ones.
We add these to the queue, and repeat the process,
until we find cells with a sufficient number $B$ of blocks.
See \cref{alg:pnas} for the pseudocode,
and \cref{fig:PNAStree} for an illustration.

\subsection{Performance Prediction with Surrogate Model}
\label{sec:surrogate}

As explained above, we need a mechanism to predict the final
performance of a cell before we actually train it.
There are at least three desired properties of such a predictor:
\begin{itemize}
    \item \textit{Handle variable-sized inputs}: We need the predictor to work for variable-length input strings.
    In particular, it should be able to predict the performance of any cell with $b+1$ blocks, even if it has only been trained on cells with up to $b$ blocks.
    
  
  \item \textit{Correlated with  true performance}: we do not necessarily need to achieve low mean squared error, but we do want the predictor to rank models in roughly the same order as their true performance values.

    \item \textit{Sample efficiency}: We want to train and evaluate as few cells as possible, which means the training data for the predictor will be scarce. 
\end{itemize}

The requirement that the predictor be able to handle variable-sized strings 
immediately suggests the use of an RNN, and indeed this is one of the methods we try.
In particular, we use an LSTM that reads a sequence of length $4b$
(representing $I_1$, $I_2$, $O_1$ and $O_2$ for each block), and the input at each step is a one-hot vector of size $|\mathcal{I}_b|$ or $|\mathcal{O}|$, followed by embedding lookup.
We use a shared embedding of dimension $D$ for the tokens $I_1, I_2 \in \mathcal{I}$,
and another shared embedding for $O_1, O_2 \in \mathcal{O}$.
The final LSTM hidden state goes through a fully-connected layer and sigmoid to regress the validation accuracy. 
We also try a simpler MLP baseline in which we convert the cell to a fixed length vector as follows:
we embed each token into an $D$-dimensional vector,
concatenate the embeddings for each block to get an $4D$-dimensional vector,
and then average over blocks.
Both models are trained using $L_1$ loss.

When training the predictor,
one approach is to
update the parameters of the predictor using the new data
using a few steps of SGD.
However, since the sample size is very small, we fit
an ensemble of 5 predictors, each fit (from scratch)
to 4/5 of all the data available at each step of the search process.
We observed empirically that this reduced the variance of the predictions.

In the future, we plan to investigate other kinds of predictors,
such as Gaussian processes with string kernels (see e.g., \cite{Baisero2015}),
which may be more sample efficient to train and produce predictions with uncertainty estimates.

%% file: results.tex
\section{Experiments and Results}
\label{sec:results}

\input{experimental-details}

\input{predictor-accuracy}

\input{efficiency}

\input{cifar}

\input{imagenet}

%% file: experimental-details.tex
\subsection{Experimental Details}

Our  experimental setting follows \cite{DBLP:journals/corr/ZophVSL17}.
In particular, we conduct most of our experiments on CIFAR-10 \cite{krizhevsky2009learning}.
CIFAR-10  has 50,000 training images and 10,000 test images.
We use 5000 images from the training set as a validation set.
All images are whitened, and $32 \times 32$ patches are cropped from images upsampled to $40 \times 40$.
Random horizontal flip is also used.
After finding a good model on CIFAR-10, we evaluate its quality on
ImageNet classification in \cref{sec:imagenet}.

For the MLP accuracy predictor, the embedding size is 100,
and we use 2 fully connected layers, each with 100 hidden units.
For the RNN accuracy predictor, we use an LSTM, 
and the hidden state size and embedding size are both 100.
The embeddings use uniform initialization in range [-0.1, 0.1].
The bias term in the final fully connected layer is initialized to 1.8 (0.86 after sigmoid) to account for the mean observed accuracy of all $b = 1$ models.
We use the Adam optimizer \cite{DBLP:journals/corr/KingmaB14} with learning rate 0.01 for the $b = 1$ level and 0.002 for all following levels.

Our training procedure for the CNNs follows the one used in
 \cite{DBLP:journals/corr/ZophVSL17}.
 During the search we evaluate $K = 256$ networks at each stage
 (136 for stage 1, since there are only 136 unique cells with 1 block),
we use a maximum cell depth of $B = 5$ blocks,
we use  $F = 24$ filters in the first convolutional cell,
we unroll the cells for $N = 2$ times,
and each child network is trained for 20 epochs using initial learning rate of 0.01 with cosine decay \cite{DBLP:journals/corr/LoshchilovH16a}.

%% file: predictor-accuracy.tex
\subsection{Performance of the Surrogate Predictors}
\label{sec:predictor}

In this section, we compare the performance of different surrogate predictors.
Note that at step $b$ of PNAS, we train the predictor
on the observed performance of cells with up to $b$ blocks,
but we apply it to cells with $b+1$ blocks.
We therefore consider predictive accuracy both for cells with sizes that have been seen before
(but which have not been trained on),
and for cells which are one block larger than the training data.

\newcommand{\loss}[1]{L(#1)}
\newcommand{\acc}[1]{A(#1)}

\begin{algorithm}[b]
\begin{algorithmic}
\FOR{$b=1:B-1$}
\FOR{$t=1:T$}
\STATE $\mathcal{S}_{b,t,1:K}$ = random sample of $K$ models from $\mathcal{U}_{b,1:R}$
\STATE $\pi_{b,t}$ = fit($\mathcal{S}_{b,t,1:K}$, $\acc{\mathcal{S}_{b,t,1:K}}$) // {\it Train or finetune predictor}
\STATE $\hat{A}_{b,t,1:K}$ = predict($\pi_{b,t}$, $\mathcal{S}_{b,t,1:K}$) // {\it Predict on same $b$}
\STATE $\tilde{A}_{b+1,t,1:R}$ = predict($\pi_{b,t}$, $\mathcal{U}_{b+1,1:R}$) // {\it Predict on next $b$}
\ENDFOR
\ENDFOR
\end{algorithmic}
\caption{Evaluating performance of a predictor  on a random dataset.
}
\label{alg:pred}
\end{algorithm}

More precisely, let $\mathcal{U}_{b,1:R}$ be a set of randomly chosen cells with $b$ blocks,
where $R=10,000$.
(For $b=1$, there are only 136 unique cells.)
We convert each of these to CNNs, and  train them for $E=20$ epochs.
(Thus in total we train $\sim (B-1) \times R = 40,000$ models for 20 epochs each.)
We now use this random dataset to evaluate the performance of the predictors
using the pseudocode in \cref{alg:pred},
where $\acc{\mathcal{H}}$ returns the true validation set accuracies of the models in some set $\mathcal{H}$.
In particular, for each size $b=1:B$, and for each trial $t=1:T$ (we use $T=20$),
we do the following:
randomly select $K=256$ models (each of size $b$)  from $\mathcal{U}_{b,1:R}$ to generate
a training set $\mathcal{S}_{b,t,1:K}$;
fit the predictor on the training set;
evaluate the predictor on the training set;
and finally 
evaluate the predictor on the set of all unseen random models of size $b+1$.

\begin{figure}[t]
\centering
\includegraphics[width=0.9\linewidth]{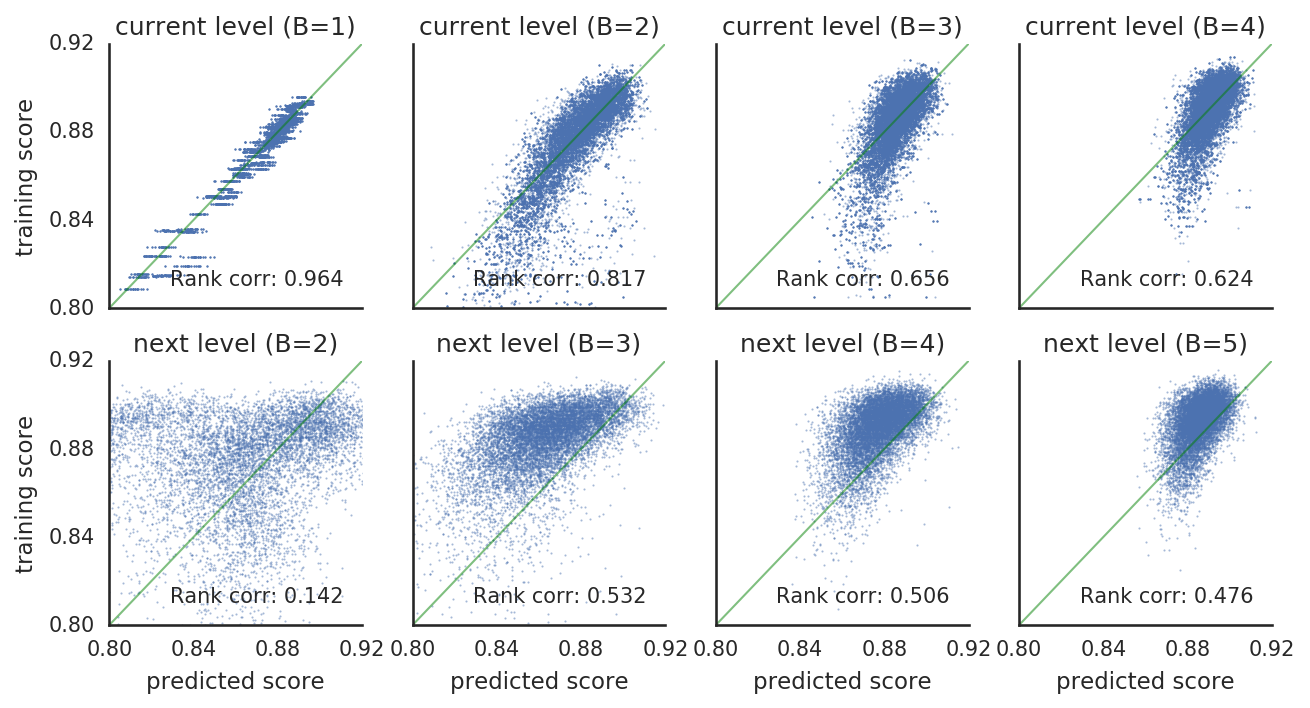}
\caption{Accuracy of MLP-ensemble predictor.
Top row: true vs predicted accuracies on models from the training set over different trials.
Bottom row:  true vs predicted accuracies on models from the set of all unseen larger models. Denoted is the mean rank correlation from individual trials.
}
\label{fig:MLP}
\end{figure}

\begin{table}[t]
\begin{center}
\begin{tabular}{l|cc|cc|cc|cc|}
\toprule
&
\multicolumn{2}{|c|}{$b=1$}
&
\multicolumn{2}{|c|}{$b=2$}
&
\multicolumn{2}{|c|}{$b=3$}
&
\multicolumn{2}{|c|}{$b=4$}
\\
Method
& $\hat{\rho}_1$ & $\tilde{\rho}_2$
& $\hat{\rho}_2$ & $\tilde{\rho}_3$
& $\hat{\rho}_3$ & $\tilde{\rho}_4$
& $\hat{\rho}_4$ & $\tilde{\rho}_5$
\\
\midrule
MLP & 0.938 & 0.113 & 0.857 & 0.450 & 0.714 & 0.469 & 0.641 & 0.444 \\
RNN & 0.970 & {\bf 0.198} & {\bf 0.996} & 0.424 & 0.693 & 0.401 & {\bf 0.787} & 0.413 \\
MLP-ensemble & 
{\bf 0.975} & 0.164 & 0.786 & {\bf 0.532} & 0.634 & {\bf 0.504} & 0.645 & {\bf 0.468} \\
RNN-ensemble &
0.972 & 0.164 & 0.906 & 0.418 & {\bf 0.801} & 0.465 & 0.579 & 0.424 \\
\bottomrule
\end{tabular}
\end{center}
\caption{Spearman rank correlations of different predictors
on the training set, $\hat{\rho}_b$,
and when extrapolating to unseen larger models,
$\tilde{\rho}_{b+1}$.
See text for details.
}
\label{tab:spearman}
\end{table}

The top row of \cref{fig:MLP} shows a scatterplot of
the true accuracies of the models in the training sets, $\acc{\mathcal{S}_{b,1:T,1:K}}$,
vs the predicted accuracies,
$\hat{A}_{b,1:T,1:K}$ (so there are $T \times K = 20 \times 256  = 5120$ points in each plot,
at least for $b>1$).
The bottom row plots 
the true accuracies on the set of larger models,
$\acc{\mathcal{U}_{b+1,1:R}}$,
vs the predicted accuracies
$\tilde{A}_{b+1,1:R}$ (so there are $R=10$K points in each plot).
We see that the predictor performs well on models from the training set,
but not so well when predicting larger models.
However, performance does increase as the predictor is trained on 
more (and larger) cells.

\cref{fig:MLP} shows the results
using an ensemble of MLPs.
The scatter plots for the other predictors look similar.
We can summarize each scatterplot using the Spearman rank correlation coefficient.
Let  $\hat{\rho}_{b}$ = rank-correlation($\hat{A}_{b,1:T,1:K}$,  $\acc{\mathcal{S}_{b,1:T,1:K}}$)
and
 $\tilde{\rho}_{b+1}$ = rank-correlation($\tilde{A}_{b+1,1:R}$,  $\acc{\mathcal{U}_{b+1,1:R}}$).
\cref{tab:spearman} summarizes these statistics across different levels.
We  see that for predicting the training set, the RNN does better than the MLP,
but for predicting the performance on unseen larger models
(which is the setting we care about in practice),
the MLP seems to do slightly better.
This will be corroborated by our end-to-end test in \cref{sec:efficiency}, and is likely due to overfitting.
We also see that for the extrapolation task, ensembling seems to help.

%% file: efficiency.tex
\subsection{Search Efficiency}
\label{sec:efficiency}

\begin{figure}[t]
    \centering
    \includegraphics[width=0.65\linewidth]{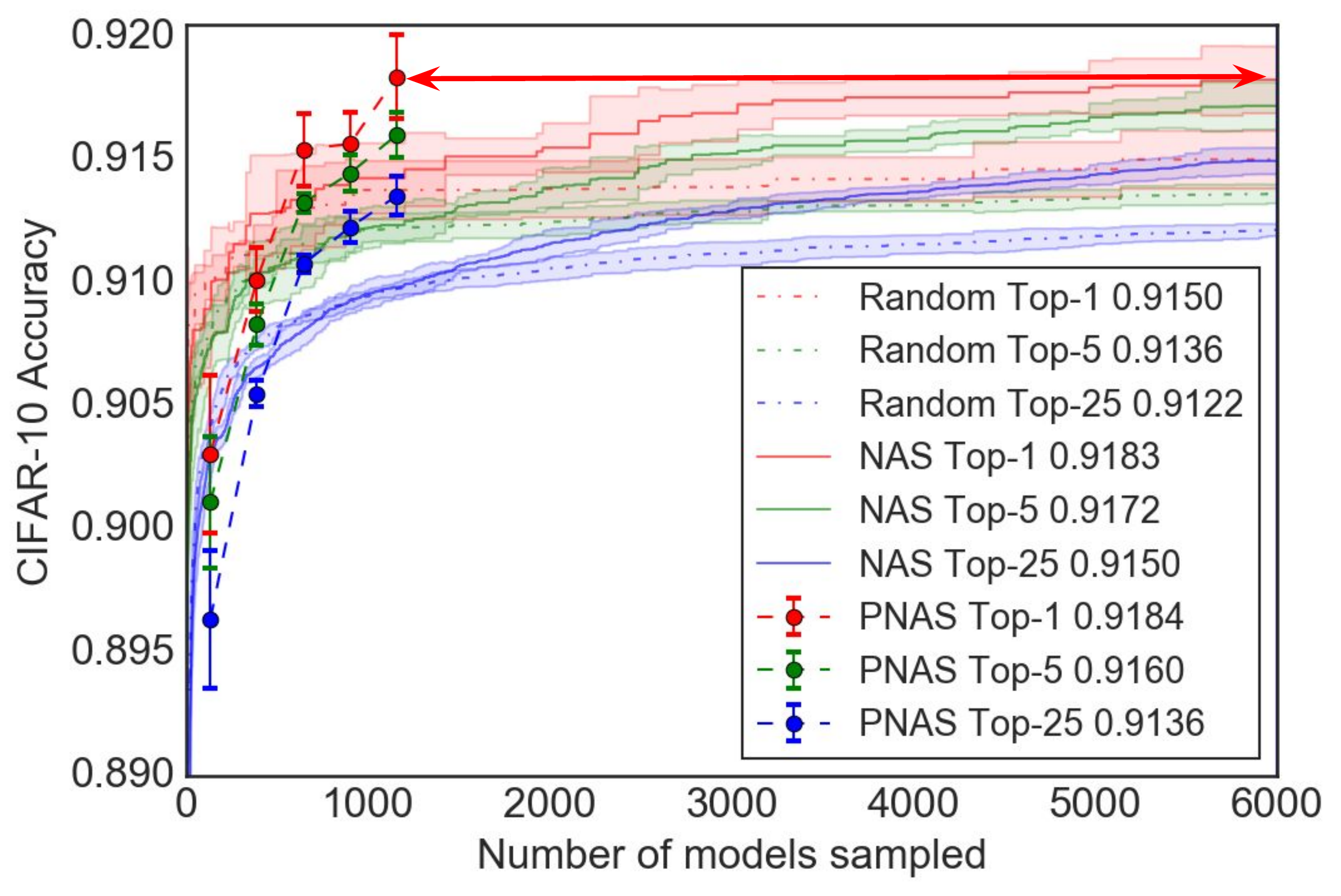}
    \caption{
    Comparing the relative efficiency of NAS,  PNAS and random search under the same search space.
    We plot mean accuracy (across 5 trials) on CIFAR-10 validation set of the top
    $M$ models, for $M \in \{1,5,25\}$, found by each method
    vs number of models which are trained and evaluated.
    Each model is trained for 20 epochs.
    Error bars and the colored regions denote standard deviation of the mean.
    }
    \label{fig:efficiency}
\end{figure}

In this section, we compare the efficiency of PNAS to two other methods:
random search and the \NAScell\ method.
To perform the comparison,
we run PNAS for $B = 5$,
and at each iteration $b$, we record the set $\mathcal{S}_b$  of $K=256$ models of size $b$
that it picks,
and evaluate them on the CIFAR-10 validation set (after training for 20 epochs each).
We then compute the validation accuracy of the top $M$ models
for $M \in \{1,5,25\}$.
To capture the variance in performance of a given model due to randomness of
the parameter initialization and optimization procedure, we repeat this process 5 times.
We plot the mean and standard error of this statistic in \cref{fig:efficiency}.
We see that the mean performance of the top $M \in \{1,5,25\}$ models steadily increases, as we search
for larger models.
Furthermore, performance is better when using an MLP-ensemble (shown in \cref{fig:efficiency})
instead of an RNN-ensemble (see supplementary material),
which is consistent with
\cref{tab:spearman}.

For our random search baseline, we uniformly sample 6000 cells of size $B=5$ blocks from
the random set of models $\mathcal{U}_{5,1:R}$
described in \cref{sec:predictor}.
 \cref{fig:efficiency} shows that PNAS significantly outperforms this baseline.

Finally, we compare to \NASRL.
Each trial sequentially searches 6000 cells of size $B=5$ blocks.
At each iteration $t$, we define $H_t$ to be the set of all cells visited so far
by the RL agent. We compute the 
validation accuracy of the top  $M$ models in $H_t$, and
plot the mean and standard error of this statistic in \cref{fig:efficiency}.
We see that the mean performance  steadily increases, but at a slower rate than PNAS.

To quantify the speedup factor compared to \NASRL,
we compute the number of models that are trained and evaluated until
the mean performance of PNAS and \NASRL\ are equal
(note that PNAS produces models of size $B$
after evaluating
 $|\mathcal{B}_1| + (B-1) \times K$ models, which is  1160 for $B=5$).
The results are shown in
\cref{tab:speedup}.
We see that PNAS is up to 5 times faster in terms of the number of models
it trains and evaluates.

\begin{table}[t]
    \begin{center}
    \begin{tabular}{l|r|c|c|c|c|c}
    \toprule
    $B$ & Top & Accuracy & $\#$\;PNAS & $\#$\;\NASRL\ & Speedup ($\#$ models) & Speedup ($\#$ examples) \\ 
    \midrule
    5 & 1 & 0.9183 & 1160 & 5808 & 5.0 & 8.2 \\
    5 & 5 & 0.9161 & 1160 & 4100 & 3.5 & 6.8 \\
    5 & 25 & 0.9136 & 1160 & 3654 & 3.2 & 6.4 \\ 
    \bottomrule
    \end{tabular}
    \end{center}
    \caption{Relative efficiency of PNAS (using MLP-ensemble predictor) and NAS under the same search space.
    $B$ is the size of the cell, 
    ``Top'' is the number of top models we pick,
    ``Accuracy'' is their average validation accuracy,
    ``$\#$ PNAS'' is the number of models evaluated by PNAS,
    ``$\#$ NAS'' is the number of models evaluated by NAS to achieve the desired accuracy.
    Speedup measured by number of examples is greater than speedup in terms of number of models, because NAS has an additional reranking stage, that trains the top 250 models for 300 epochs each before picking the best one.
    }
    \label{tab:speedup}
\end{table}

Comparing the number of models explored during architecture search is one measure of efficiency.
However, some methods, such as NAS, employ a secondary reranking stage to determine the best model; PNAS does not perform a reranking stage but uses the top model from the search directly. 
A more fair comparison is therefore to count the total number of examples processed through SGD throughout the search.
Let $M_1$ be the number of models trained during search,
and let $E_1$ be the number of examples used to train each model.\footnote{The number of examples is equal to the number of SGD steps
times the batch size.
Alternatively, it can be measured in terms of number of epoch (passes through the data),
but since different papers use different sized training sets, we avoid this measure.
In either case, we assume the number of examples is the same for every model, since none of the methods
we evaluate use early stopping.
} %
The total number of examples is therefore $M_1 E_1$.
However, for methods with the additional reranking stage,
the top $M_2$ models from the search procedure are trained using $E_2$ examples each, before returning the best.
This results in a total cost of 
$M_1 E_1 + M_2 E_2$.
For NAS and PNAS, 
 $E_1=900$K for NAS and PNAS since they use 20 epochs on a training set of size 45K.
The number of models searched to achieve equal top-1 accuracy is
$M_1=1160$ for PNAS and $M_1=5808$ for NAS.
For the second stage, NAS trains the top $M_2=250$ models for 
$E_2=300$ epochs before picking the best.\footnote{This additional stage is quite important for NAS, as the NASNet-A cell was originally ranked 70th among the top 250.
}
Thus we see  that PNAS is about 8 times faster than NAS when taking into account the total cost.

\newcommand{\MPNAS}{1160} 
\newcommand{\MNAS}{20000} 
\newcommand{\MNAStwo}{250}
\newcommand{\MEA}{7000} 

\newcommand{\EPNAS}{$9 \times 10^5$} 
\newcommand{\ENAS}{$9 \times 10^5$} 
\newcommand{\ENAStwo}{$1.35 \times 10^7$} 
\newcommand{\EEA}{$1.28 \times 10^7$} 

\newcommand{\numPNAS}{$1.04 \times 10^9$}
\newcommand{\numNAS}{$2.14 \times 10^{10}$}
\newcommand{\numEA}{$8.96 \times 10^{10}$}


\newcommand{\ShlensEPNAS}{$0.9\,$M} 
\newcommand{\ShlensENAS}{$0.9\,$M} 
\newcommand{\ShlensENAStwo}{$13.5\,$M} 
\newcommand{\ShlensEEA}{$1.28\,$M} 

\newcommand{\ShlensnumPNAS}{$1.0\,$B}
\newcommand{\ShlensnumNAS}{$21.4\,$B}
\newcommand{\ShlensnumEA}{$8.96\,$B}

%% file: cifar.tex
\subsection{Results on CIFAR-10 Image Classification}
\label{sec:cifar}

\begin{table}[t]
\begin{center}
\begin{tabular}{l c c c|c c|c c c c c}
\toprule
Model & $B$ & $N$ & $F$ & Error & Params & $M_1$ & $E_1$ & $M_2$ & $E_2$ & Cost
\\
\midrule
NASNet-A \cite{DBLP:journals/corr/ZophVSL17} & 5 & 6 & 32 & 3.41 & 3.3M & 20000 & 0.9M & 250 & 13.5M & 21.4-29.3B
\\
NASNet-B \cite{DBLP:journals/corr/ZophVSL17} & 5 & 4 & N/A & 3.73 & 2.6M & 20000 & 0.9M & 250 & 13.5M & 21.4-29.3B
\\
NASNet-C \cite{DBLP:journals/corr/ZophVSL17} & 5 & 4 & N/A  & 3.59 & 3.1M & 20000 & 0.9M & 250 & 13.5M & 21.4-29.3B
\\
\midrule
Hier-EA \cite{Liu2017} & 5 & 2 & 64 &  3.75$\pm$0.12 & 15.7M & 7000 & 5.12M & 0 & 0 & 35.8B\tablefootnote{In Hierarchical EA, the search phase trains 7K models (each for 4 times to reduce variance) for 5000 steps of batch size 256. Thus, the total computational cost is 7K $\times$ 5000 $\times$ 256 $\times$ 4 = 35.8B. }  \\
AmoebaNet-B \cite{DBLP:journals/corr/abs-1802-01548} & 5 & 6 & 36 & 3.37$\pm$0.04 & 2.8M & 27000 & 2.25M & 100 & 27M & 63.5B\tablefootnote{The total computational cost for AmoebaNet consists of an architecture search and a reranking phase. The architecture search phase trains over 27K models each for 50 epochs. Each epoch consists of 45K examples. The reranking phase searches over 100 models each trained for 600 epochs. Thus, the architecture search is 27K $\times$ 50 $\times$ 45K = 60.8B examples. The reranking phase consists of 100 $\times$ 600 $\times$ 45K = 2.7B examples. The total computational cost is 60.8B + 2.7B = 63.5B.} \\
AmoebaNet-A \cite{DBLP:journals/corr/abs-1802-01548} & 5 & 6 & 36 & 3.34$\pm$0.06 & 3.2M & 20000 & 1.13M & 100 & 27M & 25.2B\tablefootnote{The search phase trains 20K models each for 25 epochs. The rest of the computation is the same as AmoebaNet-B.} \\
\midrule
PNASNet-5 & 5 & 3 & 48  & 3.41$\pm$0.09 & 3.2M & 1160 & 0.9M & 0 & 0 & 1.0B  \\
\bottomrule
\end{tabular}
\end{center}
\caption{Performance of different CNNs on CIFAR test set.
All model comparisons employ a comparable number of parameters and exclude cutout data augmentation \cite{Cutout}. 
``Error'' is the top-1 misclassification rate on the CIFAR-10 test set.
(Error rates have the form $\mu \pm \sigma$,
where $\mu$ is the average over multiple trials and $\sigma$ is the standard deviation. In PNAS we use 15 trials.)
``Params'' is the  number of model parameters.
``Cost'' is the total number of examples processed through SGD ($M_1 E_1 + M_2 E_2$) before the architecture search terminates.
The number of filters $F$ for NASNet-\{B, C\} cannot be determined (hence N/A), and the actual $E_1$, $E_2$ may be larger than the values in this table (hence the range in cost),
according to the original authors.
}
\label{tab:resultsWithTimes}
\end{table}

We now discuss the performance of our final model,
and compare it to the results of other methods in the literature.
Let PNASNet-5 denote the best CNN we discovered on CIFAR using PNAS, also visualized in \cref{fig:network} (left).
After we have selected the cell structure,
we try various $N$ and $F$ values such that the number of model parameters is around 3M,
train them each for 300 epochs using initial learning rate of 0.025 with cosine decay,
and pick the best combination based on the validation set.
Using this best combination of $N$ and $F$, we train it for
600 epochs on the union of training set and validation set.
During training we also used auxiliary classifier located at 2/3 of the maximum depth weighted by 0.4, and drop each path with probability 0.4 for regularization.

The results are shown in \cref{tab:resultsWithTimes}.
We see that PNAS can find a model with the same accuracy as NAS,
but using 21 times less compute.
PNAS also outperforms the Hierarchical EA method of \cite{Liu2017}, while using 36 times less compute.
Though the the EA method called ``AmoebaNets''
\cite{DBLP:journals/corr/abs-1802-01548}
currently give the highest accuracies (at the time of writing),
it also requires the most compute, taking 63 times more resources than PNAS.
However, these comparisons must be taken with a grain of salt, 
since the methods are searching through different spaces.
By contrast, in \cref{sec:efficiency}, we fix the search space for NAS and PNAS,
to make the speedup comparison fair.

%% file: imagenet.tex
\subsection{Results on ImageNet Image Classification}
\label{sec:imagenet}

We further demonstrate the usefulness of our learned cell by applying it to ImageNet classification.
Our experiments reveal that CIFAR accuracy and ImageNet accuracy are strongly correlated ($\rho=0.727$; see supplementary material).

To compare the performance of PNASNet-5 
to the results in other papers,
we conduct experiments under two settings:
\begin{itemize}
    \item \textit{Mobile}: Here we restrain the representation power of the CNN.
    Input image size is $224 \times 224$, and the number of multiply-add operations is under 600M.
    \item \textit{Large}: Here we compare PNASNet-5 against the state-of-the-art models on ImageNet.
    Input image size is $331 \times 331$.
\end{itemize}

In both experiments we use RMSProp optimizer, label smoothing of 0.1, auxiliary classifier located at 2/3 of the maximum depth weighted by 0.4, weight decay of 4e-5,
and dropout of 0.5 in the final softmax layer.
In the \textit{Mobile} setting, we use distributed synchronous SGD with 50 P100 workers.
On each worker, batch size is 32, initial learning rate is 0.04, and is decayed every 2.2 epochs with rate 0.97.
In the \textit{Large} setting, we use 100 P100 workers.
On each worker, batch size is 16, initial learning rate is 0.015, and is decayed every 2.4 epochs with rate 0.97.
During training, we drop each path with probability 0.4.

The results of the \textit{Mobile} setting are summarized in \cref{tab:imagenet-mobile}.
PNASNet-5 achieves slightly better performance than NASNet-A (74.2\% top-1 accuracy for PNAS vs 74.0\% for NASNet-A).
Both methods significantly surpass
the previous state-of-the-art,
which includes the manually
designed MobileNet \cite{DBLP:journals/corr/HowardZCKWWAA17} (70.6\%)
and ShuffleNet \cite{DBLP:journals/corr/ZhangZLS17} (70.9\%).
AmoebaNet-C performs the best, but note that this is a different model than their best-performing CIFAR-10 model.
\cref{tab:imagenet-large} shows that under the \textit{Large} setting, PNASNet-5 achieves higher performance (82.9\% top-1; 96.2\% top-5) than previous state-of-the-art approaches,
including SENet \cite{DBLP:journals/corr/abs-1709-01507}, NASNet-A, and AmoebaNets under the same model capacity.

\begin{table}[t]
\begin{center}
    \begin{tabular}{lcccc}
    \toprule
         Model &  Params & Mult-Adds & Top-1 & Top-5 \\
        \midrule
        MobileNet-224 \cite{DBLP:journals/corr/HowardZCKWWAA17} & 4.2M & 569M & 70.6 & 89.5 \\
        ShuffleNet (2x) \cite{DBLP:journals/corr/ZhangZLS17} & 5M & 524M & 70.9 & 89.8 \\
        \midrule
        NASNet-A ($N=4$, $F=44$)
        \cite{DBLP:journals/corr/ZophVSL17} & 5.3M & 564M & 74.0 & 91.6 \\
        AmoebaNet-B ($N=3$, $F=62$) \cite{DBLP:journals/corr/abs-1802-01548} & 5.3M & 555M & 74.0 & 91.5 \\
        AmoebaNet-A ($N=4$, $F=50$) \cite{DBLP:journals/corr/abs-1802-01548} & 5.1M & 555M & 74.5 & 92.0 \\
        AmoebaNet-C ($N=4$, $F=50$) \cite{DBLP:journals/corr/abs-1802-01548} & 6.4M & 570M & 75.7 & 92.4 \\
        \midrule
        PNASNet-5 ($N=3$, $F=54$) & 5.1M & 588M & 74.2 & 91.9 \\
    \bottomrule
    \end{tabular}
\end{center}
\caption{ImageNet classification results in the \textit{Mobile} setting.
}
\label{tab:imagenet-mobile}
\end{table}
\begin{table}[t]
\begin{center}
    \begin{tabular}{lccccc}
    \toprule
        Model & Image Size & Params & Mult-Adds & Top-1 & Top-5 \\
        \midrule
        ResNeXt-101 (64x4d) \cite{DBLP:journals/corr/XieGDTH16} & $320 \times 320$ & 83.6M & 31.5B & 80.9 & 95.6\\
        PolyNet \cite{DBLP:journals/corr/ZhangLLL16} & $331 \times 331$ & 92M & 34.7B & 81.3 & 95.8\\
        Dual-Path-Net-131 \cite{DBLP:journals/corr/ChenLXJYF17} & $320 \times 320$ & 79.5M & 32.0B & 81.5 & 95.8\\
        Squeeze-Excite-Net \cite{DBLP:journals/corr/abs-1709-01507} & $320 \times 320$ & 145.8M & 42.3B & 82.7 & 96.2\\
        \midrule
        NASNet-A ($N=6$, $F=168$) \cite{DBLP:journals/corr/ZophVSL17} & $331 \times 331$ & 88.9M & 23.8B & 82.7 & 96.2\\
        AmoebaNet-B ($N=6$, $F=190$) \cite{DBLP:journals/corr/abs-1802-01548} & $331 \times 331$ & 84.0M & 22.3B & 82.3 & 96.1 \\
        AmoebaNet-A ($N=6$, $F=190$) \cite{DBLP:journals/corr/abs-1802-01548} & $331 \times 331$ & 86.7M & 23.1B & 82.8 & 96.1 \\
        AmoebaNet-C ($N=6$, $F=228$) \cite{DBLP:journals/corr/abs-1802-01548} & $331 \times 331$ & 155.3M & 41.1B & 83.1 & 96.3 \\
        \midrule
        PNASNet-5 ($N=4$, $F=216$) & $331 \times 331$ & 86.1M & 25.0B & 82.9 & 96.2 \\
    \bottomrule
    \end{tabular}
\end{center}
\caption{ImageNet classification results in the \textit{Large} setting.}
\label{tab:imagenet-large}
\end{table}

%% file: discussion.tex
\section{Discussion and Future Work}
\label{sec:discuss}

The main contribution of this work is to show how we can accelerate the search for good CNN structures by using progressive search through the space of increasingly complex graphs, combined with a learned prediction function to efficiently identify the most promising models to explore.
The resulting models achieve the same level of performance as previous work but with a fraction of the computational cost.

There are many possible directions for future work,
including:
the use of better surrogate predictors,
such as Gaussian processes with string kernels;
the use of model-based early stopping,
such as 
\cite{Baker2017acc},
so we can stop the training of ``unpromising'' models before reaching $E_1$ epochs;
the use of ``warm starting'', to initialize the training of a larger $b+1$-sized model
 from its smaller parent;
 the use of Bayesian optimization,
in which we use an acquisition function,
such as expected improvement or upper confidence bound,
to rank the candidate models, rather than greedily picking the top $K$
(see e.g., \cite{Snoek2012,Shahriari2016});
adaptively varying the number of models $K$ evaluated at each step (e.g., reducing it over time);
the automatic exploration of speed-accuracy tradeoffs (cf., \cite{PPPnet}),
etc.


%% file: suppl.tex
\title{Progressive Neural Architecture Search (Supplementary Material)} 

\titlerunning{Progressive Neural Architecture Search}
%
\author{Chenxi Liu\inst{1}\thanks{Work done while an intern at Google.} \and
Barret Zoph\inst{2} \and
Maxim Neumann\inst{2} \and
Jonathon Shlens\inst{2} \and
Wei Hua\inst{2} \and 
Li-Jia Li\inst{2} \and 
Li Fei-Fei\inst{2,3} \and 
Alan Yuille\inst{1} \and 
Jonathan Huang\inst{2} \and 
Kevin Murphy\inst{2}
}
%
\authorrunning{C. Liu et al.}
%

\institute{Johns Hopkins University \and
Google AI \and 
Stanford University}
\maketitle              
\appendix

\section{Search Efficiency of PNAS with RNN-ensemble}

In \cref{sec:efficiency} of the paper, we focused on the performance of MLP-ensemble as the surrogate model.
Here we provide analysis of RNN-ensemble as well.

\begin{figure}[h]
\centering
\includegraphics[width=0.65\textwidth]{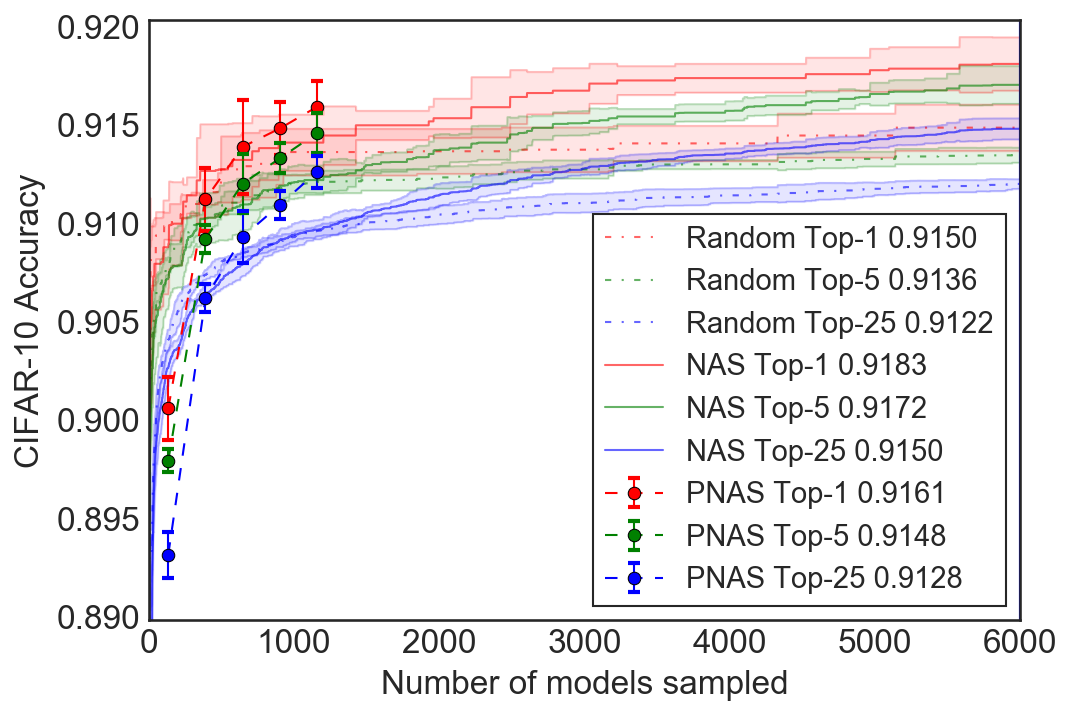}
\caption{Comparing the relative efficiency of PNAS (using RNN-ensemble) with NAS and random search under the same search space.}
\label{fig:efficiency-appendix}
\end{figure}

\begin{table}[H]
    \begin{center}
    \begin{tabular}{l|r|c|c|c|c|c}
    \toprule
    $B$ & Top & Accuracy & $\#$\;PNAS & $\#$\;NAS & Speedup ($\#$ models) & Speedup ($\#$ examples) \\ 
    \midrule
    5 & 1 & 0.9161 & 1160 & 2222 & 1.9 & 5.1 \\
    5 & 5 & 0.9148 & 1160 & 2489 & 2.1 & 5.4 \\
    5 & 25 & 0.9128 & 1160 & 2886 & 2.5 & 5.7 \\ 
    \bottomrule
    \end{tabular}
    \end{center}
    \caption{Relative efficiency of PNAS (using RNN-ensemble predictor) and NAS under the same search space.
    }
    \label{tab:speedup-appendix}
\end{table}

Again, each method is repeated 5 times to reduce the randomness in neural architecture search, and both performance mean and the variance are plotted in \cref{fig:efficiency-appendix}.
A more quantitative breakdown is given in \cref{tab:speedup-appendix}.
We see that PNAS with RNN-ensemble is about twice as efficient than NAS in terms of number of models trained and evaluated, and five times as efficient by the number of examples.
Speedup measured by number of examples is greater than speedup in terms of number of models, because NAS has an additional reranking stage, that trains the top 250 models for 300 epochs each before picking the best one.


\section{Searching Cells with More Blocks}

\begin{figure}[h]
\centering 
\includegraphics[width=0.65\textwidth]{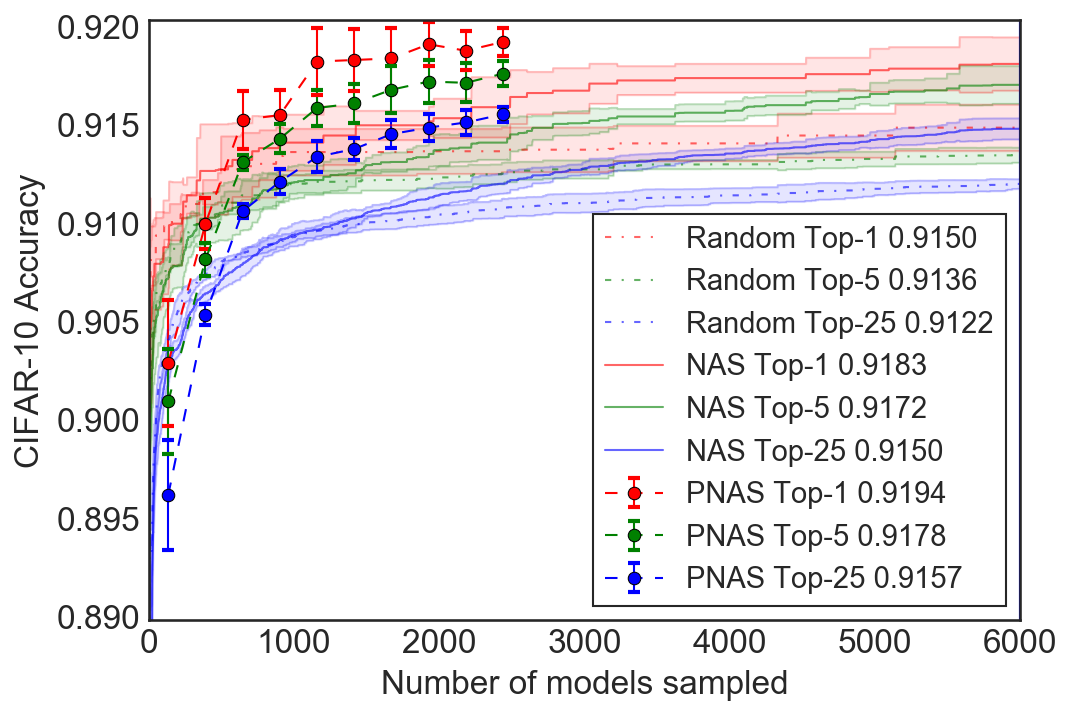}
\caption{Running PNAS (using MLP-ensemble) from cells with 1 block to cells with 10 blocks.}
\label{fig:b10}
\end{figure}

Using the MLP-ensemble predictor, we tried to continue the progressive search beyond cells with 5 blocks, all the way till $B = 10$.
The result of this experiment is visualized in \cref{fig:b10}, which extends \cref{fig:efficiency} of the main paper.
As can be seen, PNAS is able to find good performing models over the much larger search spaces of $B>5$. Note that the unconstrained search space size increases by about 4 orders of magnitude for every $B$ level, reaching $\sim 10^{33}$ possible model configurations at $B=10$. This is one of the main advantages of PNAS, to examine a highly focused search space of arbitrary size progressively.
Notice that the NAS curve for comparison is still for $B = 5$, and if we search cells with more blocks using NAS, this curve is likely to go down, because of the growth in search space.

\section{Intermediate Level PNASNet Models}

Our Progressive Neural Architecture Search algorithm explores cells from simple to complex by growing the number of blocks.
We choose $B=5$, and indeed the best model found in the final level (PNASNet-5; visualized in the left plot of \cref{fig:network}) demonstrates state-of-the-art performance.
In this section, however, we are interested in the best models found in smaller, intermediate levels, namely $b = 1, 2, 3, 4$.
We call these models PNASNet-\{1, 2, 3, 4\}.

\begin{figure}[h]
\centering
\includegraphics[height=0.36\textwidth]{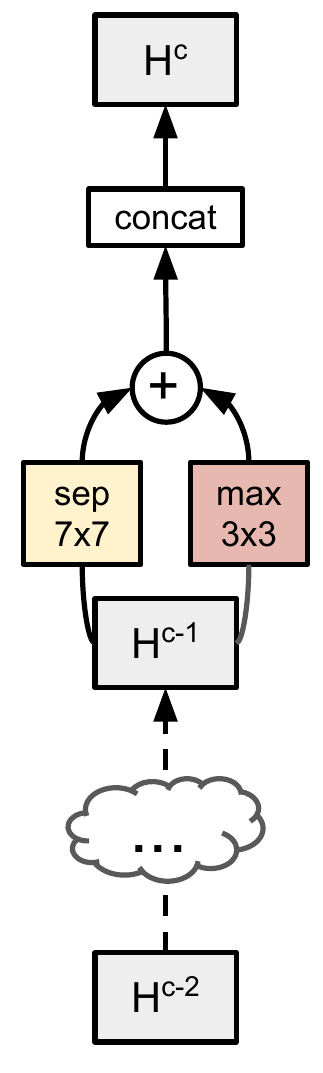}
\includegraphics[height=0.36\textwidth]{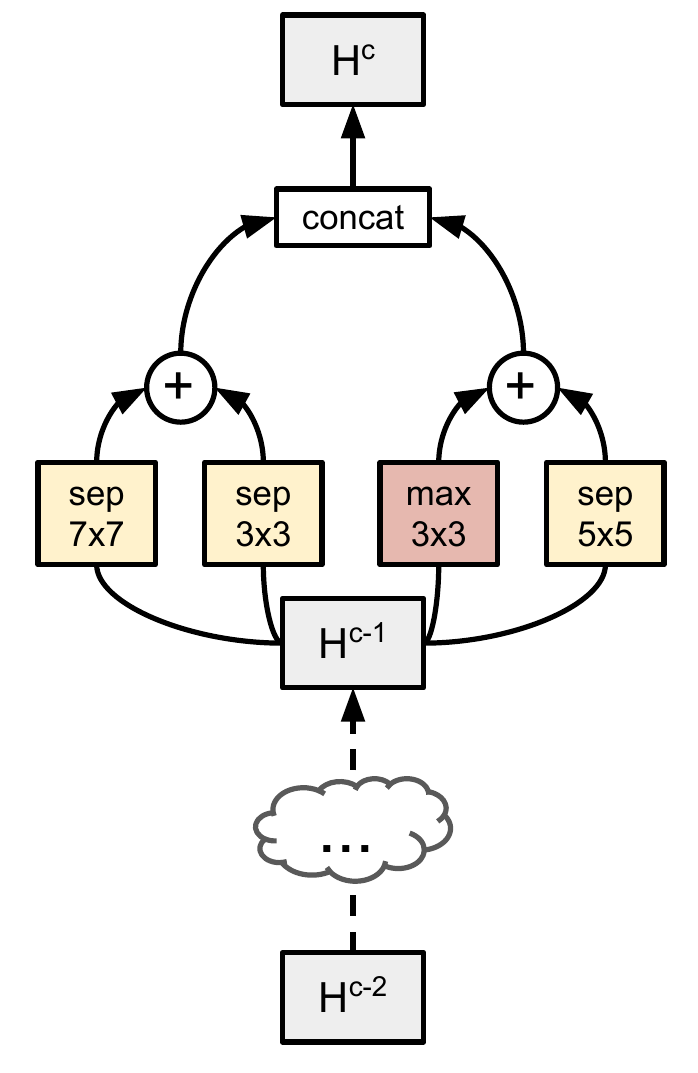}
\includegraphics[height=0.36\textwidth]{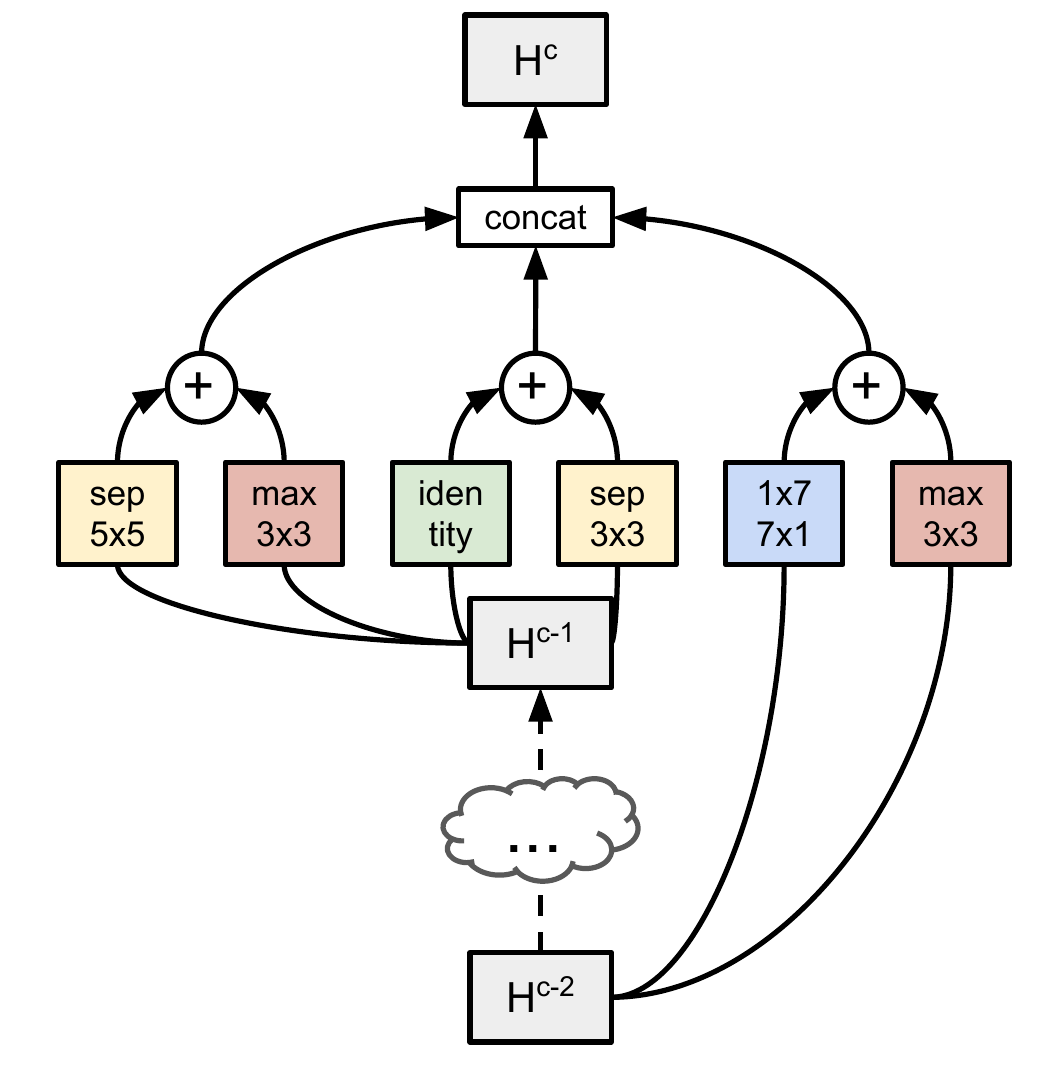}
\includegraphics[height=0.36\textwidth]{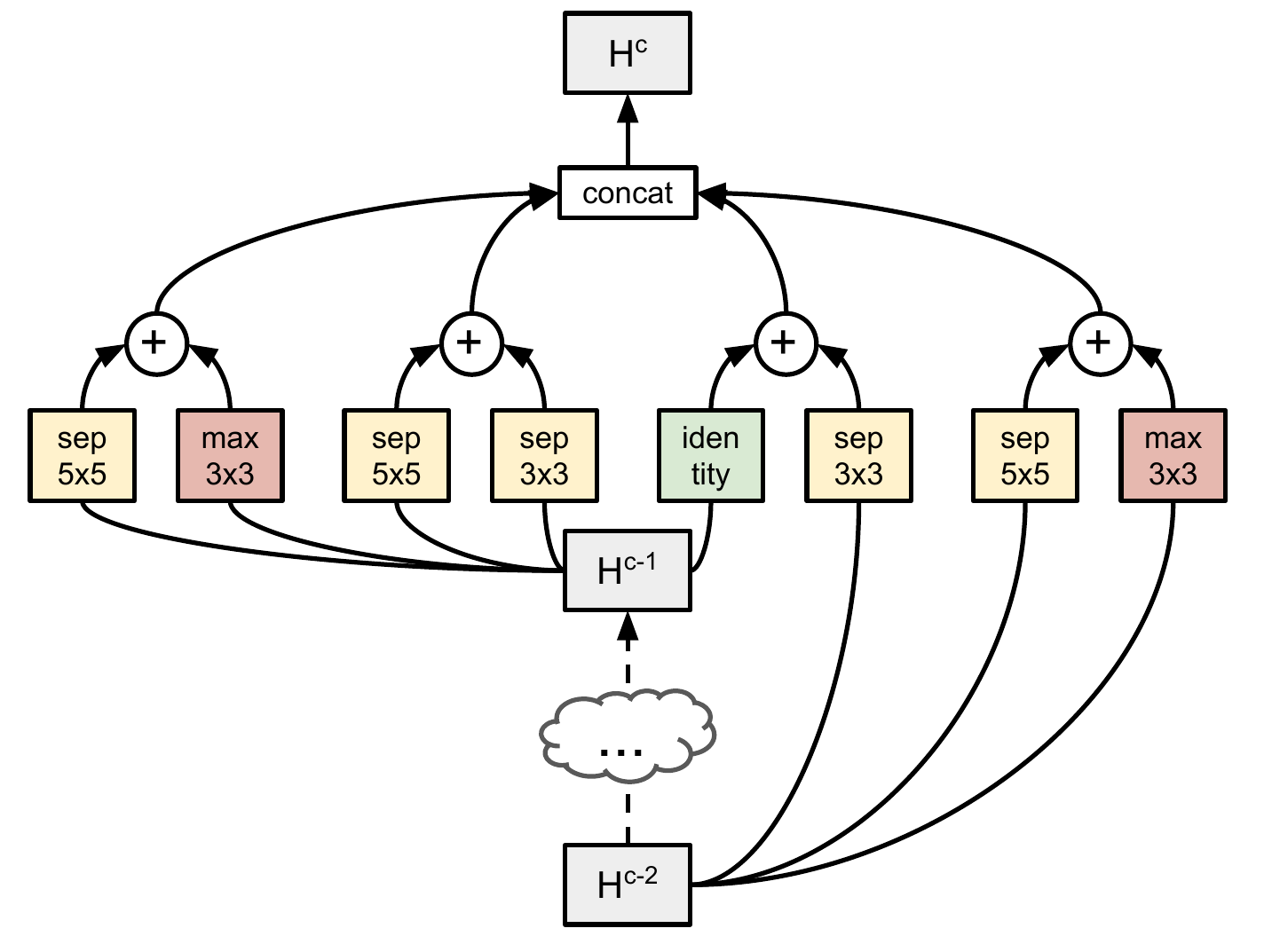}
\caption{Cell structures used in PNASNet-\{1, 2, 3, 4\}.}
\label{fig:pnasnets}
\end{figure}

\begin{table}[h!]
\begin{center}
\begin{tabular}{l c c c|c c|c c c c c}
\toprule
Model & $B$ & $N$ & $F$ & Error & Params & $M_1$ & $E_1$ & $M_2$ & $E_2$ & Cost
\\
\midrule
PNASNet-4 & 4 & 4 & 44 &  3.50$\pm$0.10 & 3.0M & 904 & 0.9M & 0 & 0 & 0.8B \\
PNASNet-3 & 3 & 6 & 32  & 3.70$\pm$0.12 & 1.8M & 648 & 0.9M & 0 & 0 & 0.6B \\
PNASNet-2 & 2 & 6 & 32 & 3.73$\pm$0.09 & 1.7M & 392 & 0.9M & 0 & 0 & 0.4B \\
PNASNet-1 & 1 & 6 & 44  & 4.01$\pm$0.11 & 1.6M & 136 & 0.9M & 0 & 0 & 0.2B \\
\bottomrule
\end{tabular}
\end{center}
\caption{Image classification performance on CIFAR test set.
``Error'' is the top-1 misclassification rate on the CIFAR-10 test set.
(Error rates have the form $\mu \pm \sigma$,
where $\mu$ is the average over multiple trials and $\sigma$ is the standard deviation. In PNAS we use 15 trials.)
``Params'' is the  number of model parameters.
``Cost'' is the total number of examples processed through SGD ($M_1 E_1 + M_2 E_2$) before the architecture search terminates.}
\label{tab:cifar}
\end{table}

We visualize their cell structures in \cref{fig:pnasnets}, and report their performances on CIFAR-10 in \cref{tab:cifar}.
We see that the test set error rate decreases as we progress from $b=1$ to $b=5$, and the performances of these PNASNets with smaller number of blocks are still competitive.

\section{Transferring from CIFAR-10 to ImageNet}

\begin{figure}[H]
    \centering
    \includegraphics[height=0.5\textwidth]{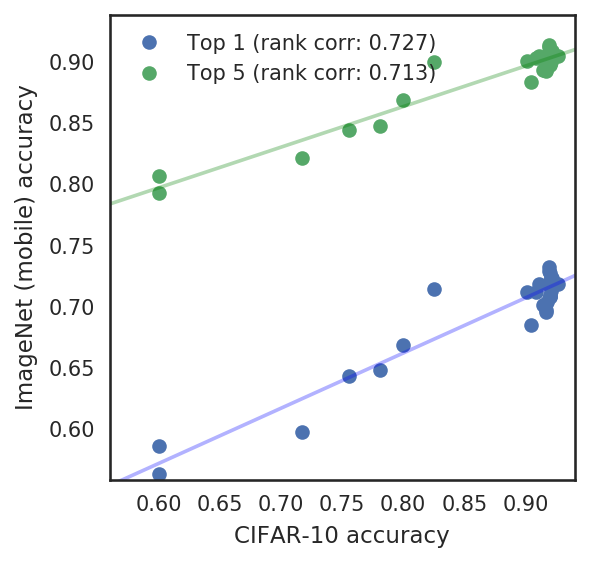}
    \caption{
    Relationship between performance on CIFAR-10 and ImageNet for different neural network architectures. The high rank correlation of 0.727 (top-1) suggests that the best architecture searched on CIFAR-10 is general and transferable to other datasets.
(Note, however, that rank correlation for the higher-value points (with CIFAR score above 0.89)
    is a bit lower: 0.505 for top-1, and 0.460 for top-5.)
    }
    \label{fig:cifar-vs-imagenet}
\end{figure}

\cref{fig:cifar-vs-imagenet} shows that the accuracy on CIFAR-10 (even for models which are only trained for 20 epochs) is strongly correlated with the accuracy on ImageNet, which proves that searching for models using CIFAR-10 accuracy as a fast proxy for ImageNet accuracy is a reasonable thing to do.

